%% file: emnlp2021.tex
\pdfoutput=1
\documentclass[11pt,a4paper]{article}
\usepackage[dvipsnames]{xcolor} 

\usepackage{authblk}
\usepackage[hyperref]{emnlp2021}

\usepackage{times}
\usepackage{latexsym}

\usepackage{microtype}

\usepackage{microtype}
\usepackage{booktabs}
\usepackage{url}
\usepackage{multirow}
\usepackage{adjustbox}

\usepackage{graphicx}
\graphicspath{ {images/} }
\usepackage{amsmath,amsfonts,amsthm,bm}
\usepackage{tabularx}
\usepackage{xspace}
\usepackage{soul}
\usepackage{xcolor, siunitx}
\usepackage{subcaption}

\newcommand{\name}{MBSR-Net\xspace}

\title{A Masked Bounding-Box Selection Based ResNet Predictor for Text Rotation Prediction}

\author{Michael Yang}
\author{Yuan Lin}
\author{ChiuMan Ho}
\affil{OPPO}
\affil{\tt \{xuewen.yang, yuan.lin, chiuman\}@oppo.com}

\date{}

\begin{document}
\maketitle
\input{0_abstract}
\input{1_introduction}

\input{3_models}

\input{4_experiments}
\input{6_conclusion}

\bibliography{eacl2021}
\bibliographystyle{acl_natbib}

\end{document}

%% file: 0_abstract.tex
\begin{abstract}

The existing Optical Character Recognition (OCR) systems are capable of recognizing images with horizontal texts.
However, when the rotation of the texts increases, it becomes harder to recognizing these texts.
The performance of the OCR systems decreases.
Thus predicting the rotations of the texts and correcting the images are important.
Previous work mainly uses traditional Computer Vision methods like Hough Transform and Deep Learning methods like Convolutional Neural Network.
However, all of these methods are prone to background noises commonly existing in general images with texts.
To tackle this problem, in this work, we introduce a new masked bounding-box selection method, that incorporating the bounding box information into the system.
By training a ResNet predictor to focus on the bounding box as the region of interest (ROI), the predictor learns to overlook the background noises. 
Evaluations on the text rotation prediction tasks show that our method improves the performance by a large margin. 

\end{abstract}

%% file: 1_introduction.tex
\section{Introduction}
\begin{figure}
    \centering
    \includegraphics[width=\columnwidth]{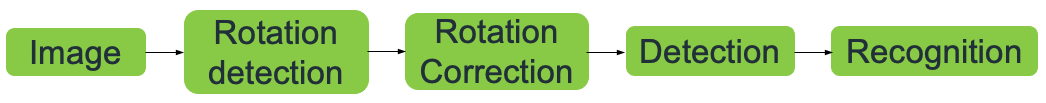}
    \caption{The OCR system in general.}
    \label{fig:process}
\end{figure}

Deep learning have emerged as the core technique in most artificial intelligent systems in the world \cite{xuewen_mm20, yang2019latent, yang-etal-2021-journalistic,Yang2020Fashion,reformer,Kaiming16}.
This owing to the ability of learning rich information from the training data through deep neural networks \cite{xuewen14,yang_license,feng2015,yang2018cross,yingru,yingruliu2}.
Optical Character Recognition (OCR) system is very helpful in many applications.
It automatically reads a printed or handwritten documents or images with texts and converts them into editable texts.
Converting images into texts to reduce storage space or transmit into other applications is very useful.
For example, UPS uses OCR systems to automatically read package information.
OCR systems can work together with Virtual Reality (VR) glasses and text-to-audio systems to help people with impaired vision to understand different scenes.
VR glasses can help people with impaired vision to capture the entire scene images.
While OCR system can extract the texts in the scene and text-to-audio system can produce audio information from texts for the people.
Developing accurate OCR system is challenging and involves several steps.
Rotation prediction and correction is done in the first step - preprocessing.
This step greatly influences the system's performance because if this system cannot be done properly, the following steps will greatly be influenced.
The whole OCR system is shown in Figure~\ref{fig:process}.

Research on text rotation prediction can be divided into two branches, the traditional Computer Vision methods based \cite{Shukla,pan,Veena,Chinara,Vijayashree} and the deep neural networks based \cite{Navneet,Akhter}.
Shukala \cite{Shukla} \textit{et al.} presented a Hough transform method to detect skews in licence plates.
Pan \textit{et al.} \cite{pan} divide the plate image into a set of $5\times 5$ non-overlapping blocks. The local orientation of each black is estimated by gradients of pixels in the block. 
Veena et al. \cite{Veena} introduced a hybrid approach with combination of Autocorrelation and Radon transform to detect the orientation and skewed of segmented number plates from vehicle image.
Chinara \textit{et al.} \cite{Chinara} proposed to use Centre of Gravity (COG) method to predict the rotation of the images.
Vijayashree \textit{et al.} \cite{Vijayashree} proposed a method to first segment the input characters and then the inclination of the character to its base is estimated using line drawing algorithm. 
Research on text rotation prediction has made great progress in recent years with the emerging technology of Convolutional Neural Networks (CNNs).
Channa \textit{et al.} \cite{Navneet} proposed to use visual attention to identify skewed vehicle license plates.
Akhter \textit{et al.} \cite{Akhter} proposed to use Convolutional Neural Networks to directly predict the rotation angles.

Though progress has been made in the text rotation prediction task, the state-of-the-art methods are far from being satisfactory.
Previous work cannot address the challenging rotation prediction problem when the background is noisy.
They usually assume that the background is clear and the images are of document or license plate types.
OCR systems of these types of images are usually of high accuracy, which limits the benefits of rotation prediction and correction.
However, in most cases, OCR systems are applied in very noisy scene images (\textit{i.g.} street-view images) or journals with a number of objects in.
Thus, the capability of focusing only on the region of interest (ROI) - the text area without being misled by the background noise is important.
In this work, we tackle rotation prediction problem by introducing bounding box information in our modeling through a masked bounding-box selection based ResNet (\name) \cite{Kaiming16} predictor.

To the best of our knowledge, our work is the first to incorporate bounding box information into the rotation prediction task.

%% file: 3_models.tex
\section{Masked Bounding-Box Selection Based ResNet Predictor}

In this section, we formally present our masked bounding-box selection based ResNet predictor (\name). We will introduce the problem definition, model architecture and the intermediate layer regularization technique.
\subsection{Problem Definition}
The rotation prediction problem can be viewed as a classification problem that given an input image with texts, the model outputs the rotation category the image belongs to.
To simplify the problem, we apply binary classification problem in this paper \footnote{We can easily extend it to multi-class classification problem without hurting the performance much}.
The two classes are `horizontal' and `vertical'.
For texts of angle $-180 \sim \ang{-135}, -45 \sim \ang{45}, 135 \sim \ang{180}$, we categorize them to the `horizontal' class. We define the rest as the `vertical' class.
Thus, the loss function can be defined as Eq. \ref{eq:loss}:
\begin{equation}
    \mathcal{L}_{cls}=-\frac{1}{N}\sum_{i=1}^{N} p(x_i)
    \label{eq:loss}
\end{equation}

\subsection{Architecture}
\begin{figure}
    \centering
    \includegraphics[width=\columnwidth]{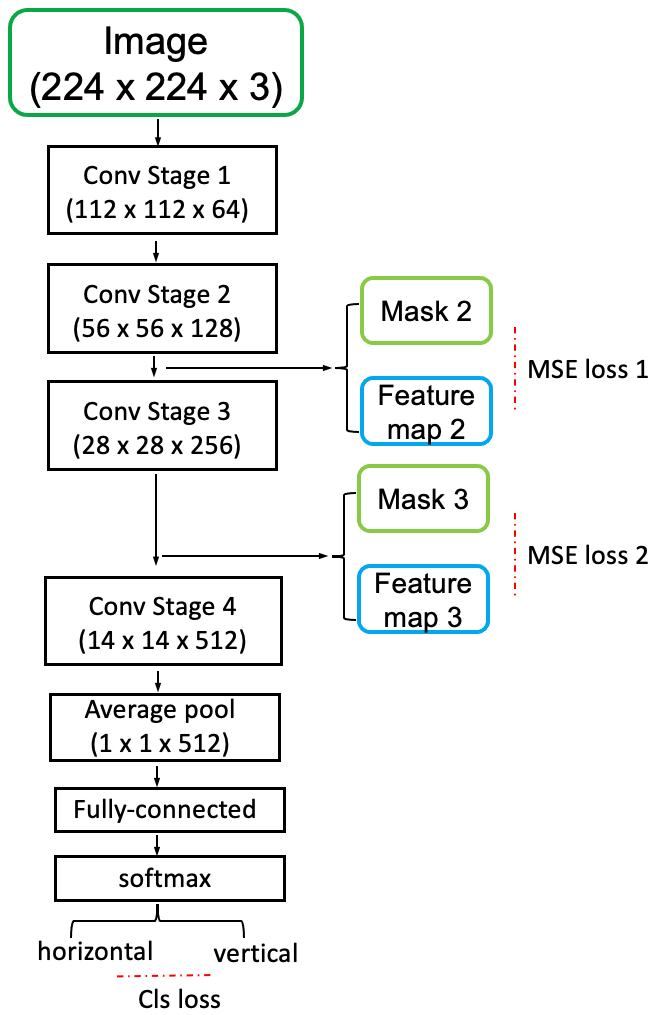}
    \caption{Our model architecture.}
    \label{fig:model}
\end{figure}

A fully convolutional network architecture based on
ResNet-18 \cite{Kaiming16} with batch normalization is adopted as our backbone. 
Our model has \textit{intermediate layer regularization} technique that is used to regularize the output features of the intermediate layers to remove the background noise (Section \ref{sec:ilr}).
The final output can be achieved after the softmax layer.
architecture is schematically illustrated in Fig. \ref{fig:model}.

\subsection{Intermediate Layer Regularization}
\label{sec:ilr}
In most cases, the background noise can greatly affect the prediction results.
To alleviate this problem, we propose an \textit{Intermediate Layer Regularization} technique that can help remove the background noise but keep only the bounding box regions or the regions of interest (ROI).
We achieve this goal by incorporating the bounding box information into the rotation prediction task.

Given an input image $\mathbf{X}$ of size $h\times w \times 3$, we first generate a mask $\mathbf{M}$ of the same size with the input image, filled with values of zero.
In the regions where there are bounding boxes, we fill with values of one.
We down-scale $\mathbf{M}$ after each Conv stage (Fig. \ref{fig:model}) to make sure that the down-scaled $\mathbf{M}_i$ is of the same size with the output features of the model $\mathbf{F}_i$.
We then regularize the output features with the down-scaled $\mathbf{M}_i$ to make them similar to each other.
Through extensive experiments, we found that we can regularize the intermediate layer output features and achieve good enough results.

Thus, we introduce a new Mean Square Error (MSE) loss as Eq. \ref{eq:loss_mse}.

\begin{equation}
    \mathcal{L}_{MSE}=\vert \vert \mathbf{F}_i - \mathbf{M}_i \vert \vert^2
    \label{eq:loss_mse}
\end{equation}

Thus, the total loss function is given as:
\begin{equation}
    \mathcal{L} = \mathcal{L}_{cls} + \mathcal{L}_{MSE}
    \label{eq:total_loss}
\end{equation}
The benefit of using intermediate layer regularization technique is that we only need the masks during train time. 
For inference, the masks are no longer needed.

%% file: 4_experiments.tex
\section{Experiments}

\subsection{Training Details}

We employ dropout with a probability 0.9 after each attention and feed-forward layer. 
Training with the overall objective function (Eq. \ref{eq:total_loss}) is done with initial learning rate of $1\times e^{-4}$ following the learning rate
scheduling strategy of \cite{Kaiming16}.

\subsection{Dataset}
We evaluate \name on a large OCR dataset containing about $100K$ image.
Each image has the corresponding bounding boxes. 
We use $80K$ images as training dataset, $10K$ images as validation dataset and $10K$ images as test dataset.

\subsection{Baselines}
We compare against two types of baselines. (i)
Conventional image processing methods \cite{Shukla,pan,Veena,Chinara,Vijayashree}. (ii) Deep neural networks based methods \cite{Navneet,Akhter}.
The results are shown in Table \ref{tab:result}.

\subsection{Results And Analysis}
\begin{table}[!t]
\centering
\begin{tabular}{ccl}
\toprule
\textbf{Model} & \textbf{Hmean}$\%$ \\
\midrule
\cite{Shukla} & 72.8 \\
\cite{pan} & 78.3  \\
\cite{Veena} & 74.5 \\
\cite{Chinara} & 77.1 \\
\cite{Vijayashree} & 76.3 \\
\cite{Navneet} & 79.2 \\
\cite{Akhter} & 81.4 \\
\name & \textbf{99.9} \\
\bottomrule
\end{tabular}

\caption{\textbf{Rotation prediction results} - shown are the Hmean scores of various models on the rotation prediction task. We highlight the \textbf{best} model in bold.}
\label{tab:result}
\end{table}

As we can tell from Table \ref{tab:result}, \name surpass all the other baselines by a large margin. 
This demonstrate the effectiveness of our \name model.

\begin{figure}
\centering
\small
\begin{subfigure}[b]{0.45\textwidth}
   \includegraphics[width=1\linewidth]{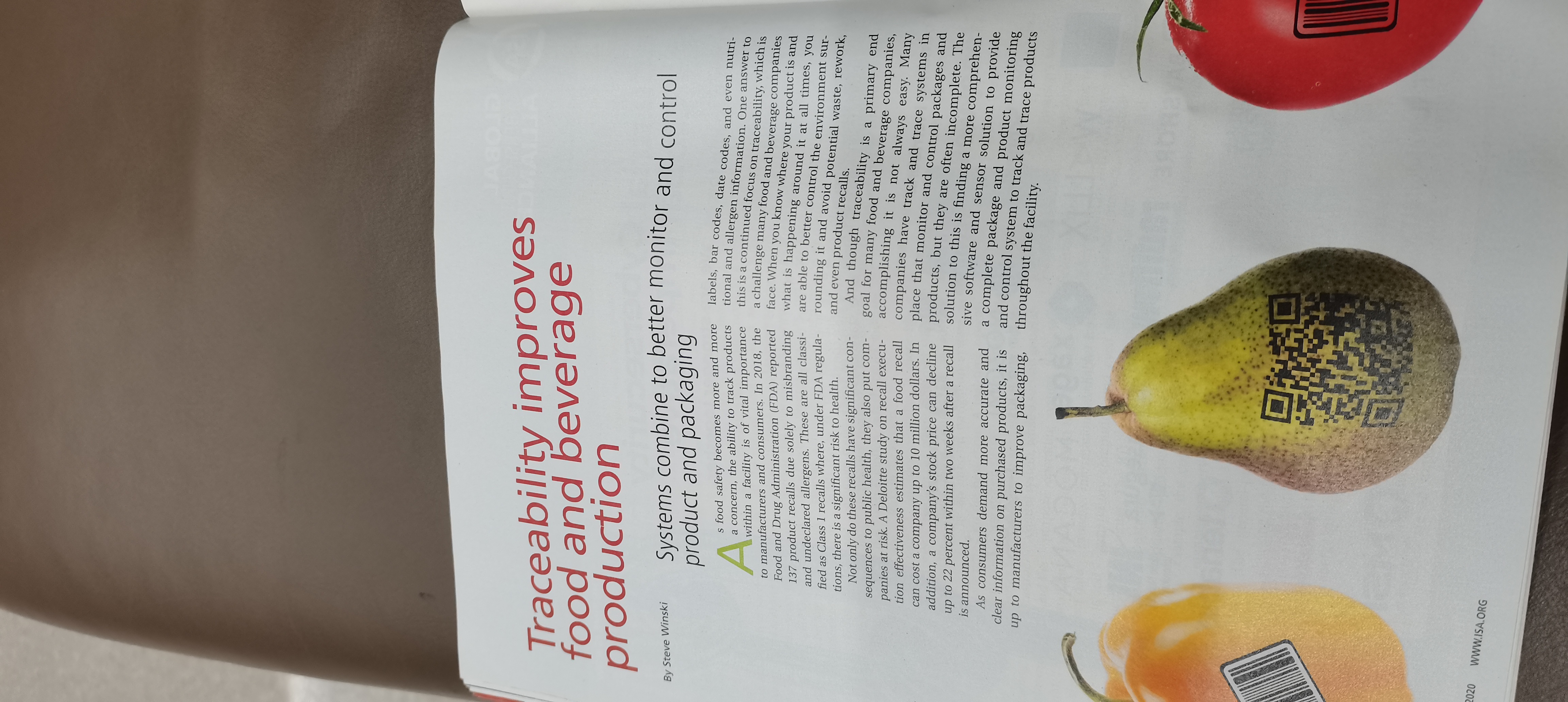}
   \caption{The original image.}
   \label{fig:ori} 
\end{subfigure}

\begin{subfigure}[b]{0.45\textwidth}
   \includegraphics[width=1\linewidth]{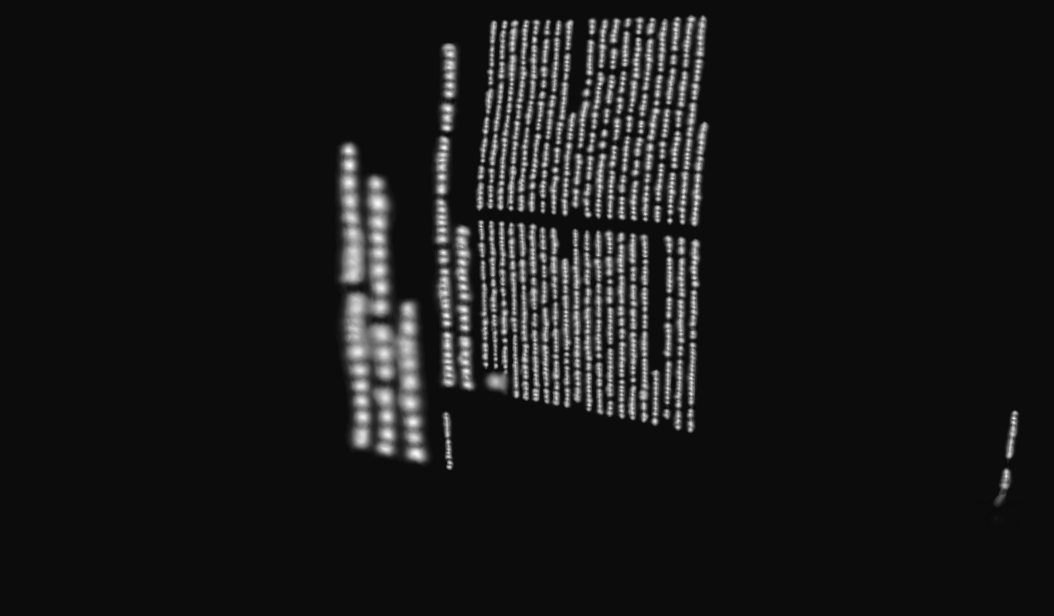}
   \caption{The final output feature map.}
   \label{fig:feature}
\end{subfigure}

\caption[]{The original image and the final output feature map.}
\label{fig:ori_fea}
\end{figure}

We showcase the original image with noisy background and the output features (Stage 4) of the image in Figure \ref{fig:ori_fea}.
We can tell that after using our method, the background becomes very clear.
And that is the reason why our method outperforms others by a large margin.

\subsection{Ablations}
In this part, we mainly focus on varying the number of intermediate layers (stages) that are regularized and see how the performance varies.

\begin{table}[!t]
\centering
\begin{tabular}{ccl}
\toprule
\textbf{Model} & \textbf{Hmean}$\%$ \\
\midrule
Stage 1 & 90.3 \\
Stage 2 & 94.5  \\
Stage 3 & 95.3 \\
Stage 4 & 94.3 \\
Stage 1,2 & 97.9 \\
Stage 2,3 & \textbf{99.9} \\
Stage 3,4 & 99.7 \\
Stage 1,2,3 & \textbf{99.9} \\
Stage 2,3,4 & \textbf{99.9} \\
Stage 1,2,3,4 & \textbf{99.9} \\
\bottomrule
\end{tabular}

\caption{\textbf{Rotation prediction results} - shown are the Hmean scores of various models on the rotation prediction task. We highlight the \textbf{best} model in bold.}
\label{tab:result2}
\end{table}

As we can tell from Table \ref{tab:result2}, with regularizing only the output features of Stage 2 and 3, we can achieve the best results.
Thus, we don't need to regularize all four stages.

%% file: 6_conclusion.tex
\section{Conclusion}
In this work, we presented a novel rotation prediction model for OCR system - \name that incorporates bounding box information into the system during the training phase.
On our rotation prediction tasks, we achieve new state-of-the-art results. 
Another key benefit of this model is that it can be easily applied to other classification tasks where removing background noise is necessary.